\documentclass{article}

\usepackage{amsmath}
\usepackage{wrapfig}
\usepackage{graphicx}
\usepackage{enumitem}
\usepackage{booktabs}
\usepackage{multirow}
\usepackage{makecell}
\usepackage[table]{xcolor}
\usepackage{boxedminipage}
\usepackage{tcolorbox}
\usepackage[colorlinks=true, urlcolor=blue]{hyperref}
\tcbuselibrary{breakable}

\definecolor{nmgray}{RGB}{229,229,229}
\definecolor{shadecolor}{RGB}{237,237,237}
\definecolor{darkpastelgreen}{rgb}{0.01, 0.75, 0.24}

\usepackage{pifont}

\usepackage{longtable}
\usepackage{adjustbox}
\usepackage{capt-of}
\usepackage{float}
\definecolor{lightblue}{rgb}{0.85,0.92,1}

\usepackage [ruled]{algorithm2e}
\definecolor{keywordcolor}{rgb}{0,0.6,0} 
\definecolor{stringcolor}{rgb}{0.58,0,0.82} 
\definecolor{commentcolor}{rgb}{0.5,0.5,0.5} 

\usepackage{fontawesome5}

\usepackage[preprint]{neurips_2025}

\usepackage[utf8]{inputenc} 
\usepackage[T1]{fontenc}    
\usepackage{hyperref}       
\usepackage{url}            
\usepackage{booktabs}       
\usepackage{amsfonts}       
\usepackage{nicefrac}       
\usepackage{microtype}      
\usepackage{xcolor}         

\title{BPO: Revisiting Preference Modeling in Direct Preference Optimization}

\author{
Lin Sun, Chuang Liu, Peng Liu, Bingyang Li, Weijia Lu, Ning Wu\\
UAES AI Lab\\ 
\texttt{\{lin.sun,chuang.liu,peng.liu2,bingyang.li,weijia.lu,ning.wu\}@uaes.com}
}

\begin{document}

\maketitle

\begin{abstract}
    Direct Preference Optimization (DPO) 
    have emerged as a popular method for aligning Large Language Models (LLMs) with human 
    preferences. While DPO effectively preserves the relative ordering between chosen and 
    rejected responses through pairwise ranking losses, it often neglects absolute reward 
    magnitudes. This oversight can decrease the likelihood of chosen responses and 
    increase the risk of generating out-of-distribution responses, leading to poor 
    performance. We term this issue \textit{Degraded Chosen Responses (DCR)}.
    To address this issue, we propose {\bf Balanced Preference Optimization (BPO)}, 
    a novel framework that dynamically balances the optimization of chosen and rejected 
    responses through two key components: \textit{balanced reward margin} and
    \textit{gap adaptor}. Unlike previous methods, BPO can fundamentally resolve DPO's DCR issue,
    without introducing additional constraints to the loss function.
    Experimental results on multiple mathematical reasoning tasks show that BPO significantly 
    outperforms DPO, improving accuracy by {\bf +10.1\%} with Llama-3.1-8B-Instruct 
    (18.8\% → 28.9\%) and {\bf +11.7\%} with Qwen2.5-Math-7B (35.0\% → 46.7\%). It also 
    surpasses DPO variants by {\bf +3.6\%} over IPO (43.1\%), {\bf +5.0\%} over SLiC (41.7\%), 
    and {\bf +3.1\%} over Cal-DPO (43.6\%) on the same model. Remarkably, our algorithm 
    requires only \textit{a single line of code modification}, making it simple to implement 
    and fully compatible with existing DPO-based frameworks. Code is available 
    at \url{https://github.com/sunlin-ai/BPO}.

\end{abstract}

\begin{figure}[ht]
  \centering
  \includegraphics[width=1.0\textwidth]{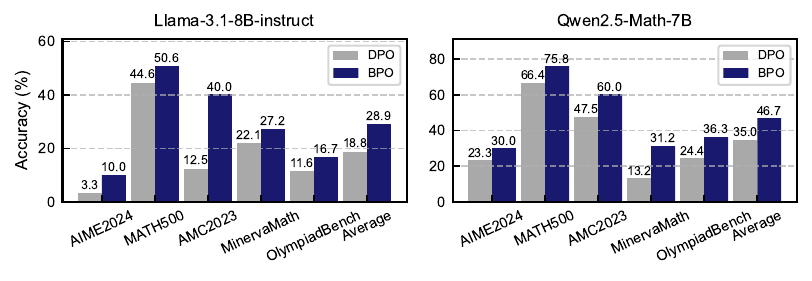}
  \vskip -0.6em
  \caption{
    Overall performance across five competition-level benchmarks (AIME2024, 
    MATH500, AMC2013, MinervaMath, and OlympiadBench). {\bf BPO} achieves an average 
    score of {\bf 28.9\%} using Llama-3.1-8B-Instruct policy generator,
    and {\bf 46.7\%} with Qwen2.5-Math-7B. This represents a substantial improvement 
    over DPO, yielding average gains of {\bf +10.1\%} and {\bf +11.7\%}, respectively.}
  \label{fig:overall}
  \end{figure}

\section{Introduction}\label{sec:intro}

Aligning Large Language Models (LLMs) with human preferences is essential to ensure their 
responses are safe, helpful, and aligned with user intent 
\cite{abs-2204-05862,Ouyang0JAWMZASR22,StiennonO0ZLVRA20}. 
While Reinforcement Learning from Human 
Feedback (RLHF) \cite{Ouyang0JAWMZASR22, ChristianoLBMLA17} has become a standard approach 
for aligning models with preferences, 
it suffers from training instability and complexity. Recent direct preference optimization 
methods \cite{XuSCTSDM024,EthayarajhXMJK24,AzarGPMRVC24,0002ZJKSLL24} 
offer a simpler alternative by replacing RLHF with supervised learning 
on preference data. These methods avoid explicit reward modeling by using policy 
likelihood to define an implicit reward, achieving both high efficiency and competitive 
performance.

\begin{figure}[!h]
    \centering
    \includegraphics[width=1.0\textwidth]{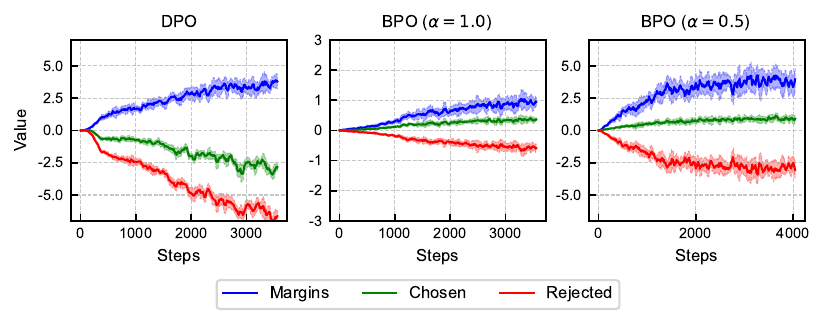}
    \vskip -0.6em
    \caption{
        In DPO, the rewards for chosen responses can drop below zero, whereas in our BPO, 
        they remain positive and continue to increase. A smaller gap adaptor $\alpha$ reduces 
        the penalty on rejected responses, while a larger $\alpha$ shifts the focus toward 
        improving chosen responses, resulting in more balanced and effective updates.
        }
    \label{fig:interaction}
\end{figure}

However, direct preference optimization methods suffer from a critical flaw: 
\textit{the likelihood of chosen responses often decreases alongside that of 
rejected responses}, as is shown in Figure \ref{fig:interaction}. We refer to 
this issue as Degraded Chosen Responses ({\bf DCR}). 
An undesirable consequence of DCR issue is that the learned policy tends to increase the 
probability of unknown out-of-distribution (OOD) responses, leading to degraded 
performance \cite{XiaoYZLH24}. We analyze the main reason for DCR issue is the 
mismatch in task difficulty: 
The shared preference modeling of direct preference optimization methods is to maximize 
the expected relative difference between the implicit rewards of chosen and 
rejected responses. This objective can be broken down into two tasks, 
increasing the probability of chosen responses and decreasing the probability 
of rejected ones. However, reducing the probability of rejected responses 
is far easier than increasing that of chosen responses, as lowering the 
likelihood of a rejected response only requires boosting arbitrary alternative 
tokens, whereas increasing  the probability of a specific response, the model 
must deeply understand the problem and identify relevant patterns to boost 
the likelihood of very specific tokens, which is an inherently harder task. 

To mitigate the DCR issue, several approaches such as DPOP \cite{abs-2402-13228} 
and Cal-DPO \cite{XiaoYZLH24} have been 
proposed. These methods aim to preserve the likelihood of chosen responses by 
introducing additional constraints into the loss function, such as enforcing 
a minimum probability threshold or regularizing the reward magnitude. While these 
modifications can help prevent the degradation of chosen response probabilities, 
they often come at a cost. The added constraints may inadvertently bias the 
optimization process, leading to suboptimal generalization and reduced model 
robustness. This lead us to the following question: 

{\centering \textit{How can we fundamentally address the DCR issue?}\par}

Our answer to this question is {\bf BPO}, a simple yet effective framework that 
addresses the DCR issue by explicitly balancing the optimization of chosen 
and rejected responses. The key intuition behind BPO is quite simple: instead of 
only maximizing the relative reward gap between chosen and rejected responses, 
we should also ensure that the absolute chosen reward is 
preserved. This can be achieved through a simple modification to existing methods. 
For instance, BPO can be implemented on top of DPO with just a single line of 
code, by replacing the relative reward margin term with a balanced reward margin. 
Moreover, BPO can easily generalize to other preference optimization functions 
(see Section \ref{thm:various_loss}).

We summarize our contributions as follows:

\begin{itemize}
    \item We propose BPO, a novel framework which 
    effectively mitigates the DCR issue by explicitly preserving the absolute likelihood 
    of chosen responses while still maximizing the reward gap between chosen and 
    rejected responses.
    \item Our method is simple, general, and easy to implement, requiring only a minor 
    modification to existing preference optimization algorithms. 
    \item Through extensive experiments and ablation studies, we demonstrate that BPO 
    consistently outperforms DPO and its variants. We present a new perspective on reparametrizing 
    the reward margin, which significantly enhances training stability and overall performance.
    \end{itemize}

\section{Methodology}
\subsection{Balanced Preference Optimization}


{\bf Problem Setup.} Let $ \mathbf{x} = [x_1, x_2, \ldots] $ denote an input sequence, 
and let $ \mathbf{y}_w = [y_1, y_2, \ldots] $ and $ \mathbf{y}_l = [y_1, y_2, \ldots] $ 
be two responses sampled from a reference language model 
$ \pi_{\text{ref}}(\mathbf{y} \vert \mathbf{x}) $. These response pairs are then presented 
to human or model-based annotators, which provide preference labels of the form 
$ \mathbf{y}_w \succ \mathbf{y}_l \mid \mathbf{x} $, indicating that $ \mathbf{y}_w $ is 
preferred over $ \mathbf{y}_l $ given the input $ \mathbf{x} $. The preference distribution 
is typically modeled using a latent reward function, as follows:

\begin{equation}
p(\mathbf{y}_w \succ \mathbf{y}_l \mid \mathbf{x}) = f\left( \beta \log \frac{\pi_\theta(\mathbf{y}_w | \mathbf{x})}{\pi_{\text{ref}}(\mathbf{y}_w | \mathbf{x})} - \beta \log \frac{\pi_\theta(\mathbf{y}_l | \mathbf{x})}{\pi_{\text{ref}}(\mathbf{y}_l | \mathbf{x})} \right).
\end{equation}

where $ f : \mathbb{R} \to [0, 1] $ is preference optimization function, which converts 
reward differences into winning probabilities. 
When $ f $ is the logistic log function, we get 
the Bradley-Terry (BT) preference model [41]. Given a dataset 
$ \mathcal{D} = \{ (\mathbf{x}^{(i)}, \mathbf{y}_w^{(i)}, \mathbf{y}_l^{(i)}) \}_{i=1}^N $ 
of human preferences, the learning objective is to optimize the policy $ \pi_\theta $ 
such that it aligns with the preference distribution while maintaining a controlled 
divergence from $\pi_{\text{ref}}$.

{\bf Balanced Reward Margin.} Standard direct preference optimization methods optimize the 
relative reward margin defined as: 

\begin{equation}
\rho^d_{\theta} = r_w - r_l.
\end{equation}

Where,
$r_w = \beta \log \pi_\theta(\mathbf{y}_w | \mathbf{x}) - \beta\log \pi_{\text{ref}}(\mathbf{y}_w | \mathbf{x})$, 
$r_l = \beta \log \pi_\theta(\mathbf{y}_l | \mathbf{x})- \beta\log \pi_{\text{ref}}(\mathbf{y}_l | \mathbf{x})$.
While maximizing $ \rho^d_\theta $ encourages $ \pi_\theta(\mathbf{y}_w | \mathbf{x}) \gg \pi_\theta(\mathbf{y}_l | \mathbf{x}) $, 
it disregards the absolute magnitudes of $ r_w $ and $ r_l $. This permits two failure 
modes: (1) Degraded Chosen Responses:  
$ \pi_\theta(\mathbf{y}_w | \mathbf{x}) $ can diminish as long as 
$ \pi_\theta(\mathbf{y}_l | \mathbf{x}) $ degrades faster, since $ \rho^d_\theta $ 
depends only on their relative difference. (2) Overestimated Rejected Responses:  
Insufficient suppression of $ \pi_\theta(\mathbf{y}_l | \mathbf{x}) $ (small $ |r_l| $) 
may lead to low-quality or out-of-distribution $ \mathbf{y}_l $. 
To address these issues, we propose a balanced reward margin:

\begin{equation}
\rho^b_{\theta} = \min\left(r_w, -r_l\right).
\end{equation}

It dynamically prioritizes the weaker component of the reward pair:
(1) when $ r_w \leq -r_l $, focuses on improving $ r_w $, ensuring $ \pi_\theta(\mathbf{y}_w | \mathbf{x}) $ aligns with high-quality responses.
(2) when $ -r_l \leq r_w $, prioritizes suppressing $ \pi_\theta(\mathbf{y}_l | \mathbf{x}) $, mitigating overconfidence in undesired outputs.
By optimizing $ \rho^b_\theta $, the policy maintains a balance between reinforcing chosen responses and penalizing rejected ones, preventing 
pathological optimization trajectories inherent to standard DPO.

{\bf Gap Adaptor.} To provide finer control over the balance, 
we introduce a gap adaptor $ \alpha \in (0,1] $ that controls the relative gap 
between $r_w $ and $r_l$. The balanced reward margin becomes:

\begin{equation}
\rho^b_\theta = \min(r_w, -\alpha r_l).
\end{equation}

The gap adaptor $ \alpha $ allows us to adjust the relative importance of suppressing 
rejected responses compared to improving chosen responses. A smaller $ \alpha $
decreases the penalty on rejected responses, while a larger $ \alpha $ places more 
emphasis on improving chosen responses, resulting in more balanced updates, as 
illustrated in Figure~\ref{fig:interaction}.

{\bf Final Objective.} Incorporating the balanced reward margin and the gap adaptor, the final loss function of BPO is defined as:

\begin{equation}
\mathcal{L}(\theta) = -\mathbb{E}_{(\mathbf{x}, \mathbf{y}_w, \mathbf{y}_l) \sim \mathcal{D}} \left[ f \left( \min \left( \beta \log \frac{\pi_\theta(\mathbf{y}_w | \mathbf{x})}{\pi_{\text{ref}}(\mathbf{y}_w | \mathbf{x})}, -\alpha \beta \log \frac{\pi_\theta(\mathbf{y}_l | \mathbf{x})}{\pi_{\text{ref}}(\mathbf{y}_l | \mathbf{x})} \right) \right) \right].
\end{equation}

By optimizing this loss, we ensure that the policy does not overfit to either the 
chosen or rejected responses, thereby maintaining a balanced and robust 
alignment with human preferences.


\subsection{Theoretical Analysis}
\label{sec:theoretical_analysis}
{\bf Gradient analysis.} When employing logistic log as preference optimization function,
which is same with standard DPO, the gradient of the BDO loss is given by (derived from the Appendix \ref{sec:gradient}):

\begin{equation}
\nabla_\theta \mathcal{L}(\theta) =
-\mathbb{E}_{(\mathbf{x}, \mathbf{y}_w, \mathbf{y}_l) \sim \mathcal{D}}
\begin{cases}
\sigma(-\beta r_w) \cdot \beta \cdot \nabla_\theta \log \pi_\theta(\mathbf{y}_w | \mathbf{x}), & \text{if } r_w < -\alpha r_l,\\
\sigma(\alpha \beta r_l) \cdot (-\alpha \beta) \cdot \nabla_\theta \log \pi_\theta(\mathbf{y}_l | \mathbf{x}), & \text{otherwise}.
\end{cases}
\end{equation}

In contrast, the gradient for DPO is:

\begin{equation}
\nabla_\theta \mathcal{L}_{\text{DPO}} = -\beta \mathbb{E}_{(\mathbf{x}, \mathbf{y}_w, \mathbf{y}_l) \sim \mathcal{D}} \left[ \sigma(-\beta (r_w - r_l)) \cdot (\nabla_\theta \log \pi_\theta(\mathbf{y}_w | \mathbf{x}) - \nabla_\theta \log \pi_\theta(\mathbf{y}_l | \mathbf{x})) \right].
\end{equation}

From this comparison, we can highlight several key advantages of BPO:

(1) Dynamic Balancing of Chosen and Rejected Response Updates.
In DPO, gradient updates are weighted based on the relative log-probabilities 
of chosen and rejected responses. This approach can lead to an overemphasis on rejected 
responses due to mismatch in task difficulty, potentially neglecting updates from chosen ones. 
In contrast, BPO 
introduces a threshold parameter $\alpha$ to dynamically adjust the contribution of 
each response type. When $ r_w < -\alpha r_l$, only the chosen response 
influences the gradient. Otherwise, the model prioritizes updating the rejected response 
via $-\alpha \beta \nabla_\theta \log \pi_\theta(\mathbf{y}_l | \mathbf{x})$, it ensures balanced learning.

(2) Accelerated Convergence.
We visualize the gradient distributions of the loss for both the relative reward margin 
$r_w - r_l$ and the balanced reward margin $\min(r_w, -r_l)$. The visualization 
reveals that when using the relative reward margin, more than half of the gradients in 
the lower-right region have very low values. This indicates that when the model parameters 
converge to this area, the updates will be exceedingly slow. In contrast, when using the 
balanced reward margin, the region with low gradient values is significantly reduced. 
As a result, the model can benefit from more substantial updates, thereby accelerating 
the convergence.

(3) Reduced Computational Overhead.
DPO requires computing gradients for both chosen and rejected responses, 
along with their probability ratios, which can be computationally intensive, especially 
on large datasets. BPO reduces computational overhead by computing gradients 
only for the response under strong preference conditions 
($\beta r_w < -\alpha \beta r_l$). This effectively halves the computational cost.

\textbf{Theorem 1.} \textit{Let $ \gamma $ be the maximized margin in the balanced reward margin constraint $ \rho^b_\theta \geq \gamma $. Under BPO, the likelihood of the chosen response satisfies:}

\begin{equation}
\pi_\theta(\mathbf{y}_w \mid \mathbf{x}) \geq \exp\left(\frac{\gamma}{\beta}\right) \pi_{\text{ref}}(\mathbf{y}_w \mid \mathbf{x}).
\end{equation}

Proof provided in Appendix~\ref{sec:proof_theorem1}, this inequality ensures that the learned policy $ \pi_\theta $ 
assigns a probability to the chosen 
response $ \mathbf{y}_w $ that is at least exponentially greater, scaled by $ \gamma / \beta $ 
than that assigned by the reference policy $ \pi_{\text{ref}} $. Importantly, this constraint 
prevents the degradation of $ \pi_\theta(\mathbf{y}_w \mid \mathbf{x}) $, as the likelihood 
ratio cannot fall below $ \exp(\gamma / \beta) $, even if $ \pi_\theta(\mathbf{y}_l \mid \mathbf{x}) $ 
is further suppressed. 

In contrast, DPO optimizes the relative margin $ r_w - r_l $, which allows 
$ \pi_\theta(\mathbf{y}_w \mid \mathbf{x}) $ to decrease as long as 
$ \pi_\theta(\mathbf{y}_l \mid \mathbf{x}) $ degrades faster. 
This can lead to a collapse in the probability of high-quality chosen responses. 

\begin{figure}[!h]
    \centering
    \vskip -1em
    \includegraphics[width=1.0\textwidth]{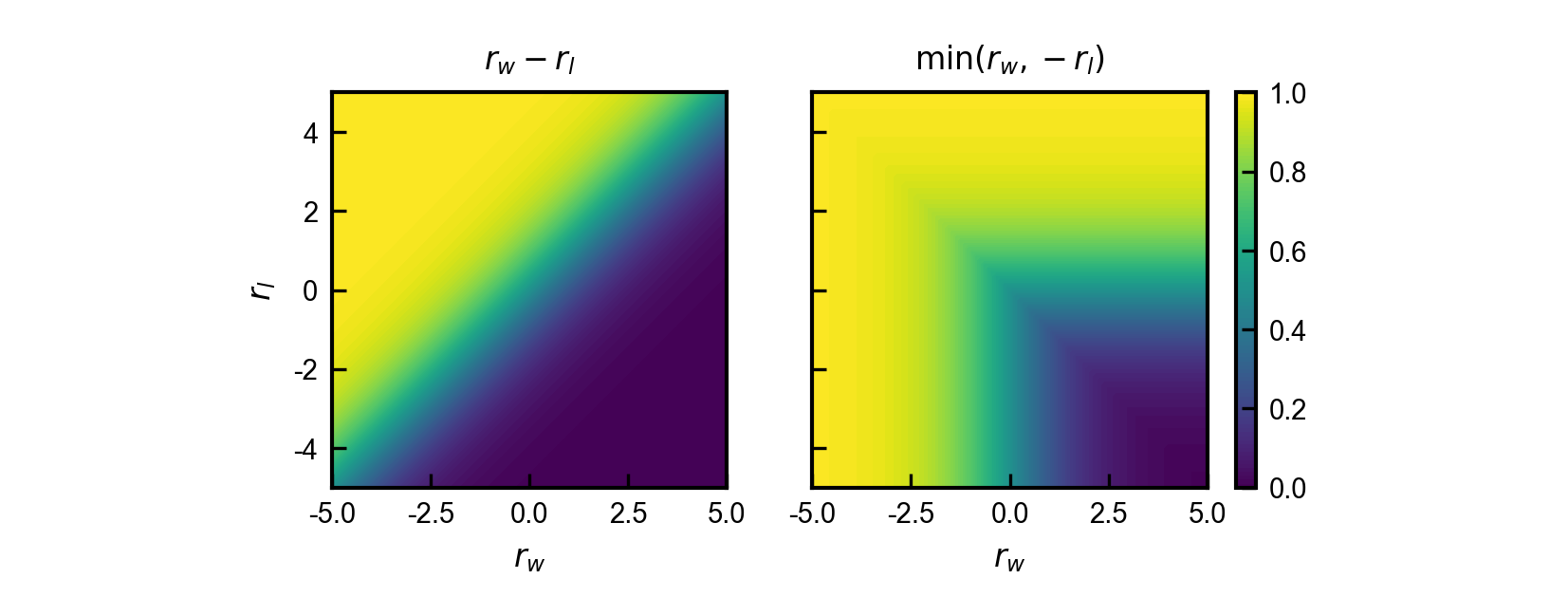}
    \vskip -0.6em
    \caption{
        The gradient distributions under the logistic log preference optimization function are 
        compared for both the relative reward margin $r_w - r_l$ and the balanced reward 
        margin $\min(r_w, -r_l)$. Notably, the region with low gradient values is significantly 
        reduced when using the balanced reward margin.
    }
    \label{fig:graident}
\end{figure}

\section{Experiment}
\label{sec:experiment}

\subsection{Experiment Setup}
\textbf{Training Dataset.} Our experiments were performed using the publicly available Open Reasoner 
Zero 57k dataset \cite{abs-2503-24290}, which consists of mathematical problems sourced from 
a variety of collections, such as AIME, MATH, Numina-Math Collection, Tulu3 MATH, OpenR1-Math-220k, 
and other open-source datasets.

\textbf{Evaluation} To comprehensively evaluate our model's reasoning capabilities, we conduct 
experiments on a diverse range of mathematical reasoning benchmarks, including AIME2024, 
MATH500 \cite{HendrycksBKABTS21}, AMC2013 \cite{LewkowyczADDMRS22}, 
MinervaMath \cite{LewkowyczADDMRS22} 
and OlympiadBench \cite{HeLBHTSHHHZLQL024}.
We report the average accuracy across these five datasets as our primary evaluation metric. 
For each dataset, we compute the pass@1 score.
To further assess the generalizability of our approach, we test BPO using two model families: 
the Llama series \cite{abs-2407-21783} and the Qwen2.5 series \cite{abs-2412-15115}. 
Specifically, we use Llama-3.1-8B-Instruct
, along with three models from the Qwen2.5 series: Qwen2.5-Math-1.5B, 
Qwen2.5-Math-7B, and Qwen2.5-Math-7B-Instruct.

\textbf{Baselines.} We compare BPO with several strong baselines. Specifically, 
we contrast BPO against GPT-4o \cite{abs-2410-21276}, as well as open-source instruction-tuned 
models such as Llama-3.1-70B-Instruct and Qwen2.5-Math-7B-Instruct.
Additionally, we compare BPO with fine-tuned models including Qwen2.5-Math-7B-Base-SFT \cite{zhang2025dpor1}, which uses 
Supervised Fine-Tuning (SFT), and Qwen2.5-7B-RAFT-Zero \cite{zhang2025dpor1}, which employs Reward-Ranked Fine-Tuning.
Furthermore, we benchmark BPO against several preference optimization methods: 
DPO \cite{RafailovSMMEF23}, IPO \cite{AzarGPMRVC24}, SLiC \cite{0002ZJKSLL24}, Cal-DPO \cite{XiaoYZLH24}, 
and DPOP \cite{abs-2402-13228}.

\textbf{Implementation Details.} Our base models is Qwen2.5-Math-7B-Base, following prior 
work \cite{zhang2025dpor1}, we sample 8 
responses per prompt and rank them based on the correctness of the final outcome. We then 
apply the max-min strategy to select 
preference pairs, specifically the response with the highest reward is paired with the one with the lowest reward. 
If all responses receive the same reward, we discard the prompt and proceed to the next one.
We train the model with a learning rate of 5e-6, a batch size of 8, for 2 epochs, using 
a maximum sequence length of 2048. The experiments are run on 2 Nvidia A40 GPUs with
BF16 precision. The prompt used by the policy generator is provided in Appendix \ref{sec:prompt}.

\subsection{Overall Performance}

\textbf{BPO achieves superior performance, outperforming DPO and its variants.} 
As shown in Table~\ref{tab:overall_performance}, the proposed method, BPO, demonstrates 
superior performance across five competition-level mathematical reasoning benchmarks , 
achieving an average accuracy of 46.7\%, significantly outperforming DPO and 
its variants. Notably, BPO excels on the challenging AIME 2024 benchmark with 30.0\% 
accuracy, showcasing its effectiveness in complex reasoning tasks. Compared to base 
models and those fine-tuned via Supervised Fine-Tuning, 
BPO shows clear advantages, highlighting the benefits of preference-based training. 
Importantly, BPO achieves these improvements through minimal 
modifications to existing frameworks, making it a practical and effective enhancement 
for large language models.

\begin{table}[!h]
\centering
\caption{Overall performance across five competition-level math reasoning benchmarks. 
Results for BPO are \colorbox{cyan!10}{shaded}. \textbf{Avg.} indicates the mean 
accuracy across all datasets. The top results are shown in \textbf{bold}. 
The table demonstrates that BPO outperforms both standard DPO and its variants, 
achieving the highest average accuracy.
}
\resizebox{1.0\textwidth}{!}{
\begin{tabular}{lcccccc}
\toprule
\multirow{2}{*}{\textbf{Method ($\downarrow$)} $\slash$ \textbf{Dataset ($\rightarrow$)}} & \multirow{2}{*}{\textbf{AIME2024}} & \multirow{2}{*}{\textbf{MATH500}} & \multirow{2}{*}{\textbf{AMC2023}} & \multirow{2}{*}{\begin{tabular}[c]{@{}c@{}} \bf Minerva \\ \bf Math \end{tabular}} & \multirow{2}{*}{\begin{tabular}[c]{@{}c@{}} \bf Olympiad \\ \bf Bench \end{tabular}} & \multirow{2}{*}{\textbf{Avg.}} \\
& \\
\midrule
GPT-4o                         & 9.3 & 76.4 & 45.8 & 36.8 & 43.3 & 43.3 \\
Llama-3.1-70B-Instruct         & 16.7 & 64.6 & 30.1 & 35.3 & 31.9 & 35.7 \\
Qwen2.5-Math-7B-Base           & 23.3 & 66.4 & 47.5 & 13.2 & 24.4 & 35.0 \\
Qwen2.5-Math-7B-Base-SFT       & 20.0 & 73.2 & 62.5 & 30.5 & 35.6 & 44.4 \\
Qwen2.5-Math-7B-Instruct       & 13.3 & 79.8 & 50.6 & 34.6 & 40.7 & 43.8  \\
Qwen2.5-7B-RAFT-Zero           & 20.0 & 77.6 & 55.0 & 30.5 & 38.7 & 44.4 \\
\midrule
DPO          & 6.7 & 71.2 & 55.0 & 39.3 & 32.9 & 41.0 \\
IPO          & 10.0 & 75.6 & 52.5 & 39.7 & 37.6 & 43.1 \\
SLiC         & 10.0 & 73.2 & 55.0 & 37.5 & 33.0 & 41.7 \\
Cal-DPO      & 20.0 & 75.4 & 62.5 & 24.3 & 35.9 & 43.6 \\
DPOP         & 23.3 & 77.0 & 57.5 & 30.9 & 35.9 & 44.9 \\
\midrule
\rowcolor{cyan!10} BPO (ours) & 30.0 & 75.8 & 60.0 & 31.2 & 36.3& {\bf 46.7} \\
\bottomrule
\end{tabular}
}
\label{tab:overall_performance}
\end{table}


\subsection{Ablation Study}

\textbf{BPO consistently outperforms DPO across all model architectures and scales.} 
The results in Table~\ref{tab:backbone} demonstrate that BPO improves performance 
across different LLM architectures and scales. Specifically, when applied to diverse 
model architectures, BPO significantly boosts average accuracy on Llama-3.1-8B-Instruct 
(from 18.8\% to 28.9\%) and Qwen2.5-Math-7B-Instruct (from 42.8\% to 48.8\%). In terms 
of model scale, BPO consistently enhances performance across parameter sizes: from 
28.2\% to 38.3\% on the smaller Qwen2.5-Math-1.5B-Base, and from 41.0\% to 46.7\% on 
the larger Qwen2.5-Math-7B-Base. Furthermore, BPO not only surpasses the closed-source 
model GPT-4o, but also outperforms instruction-tuned models such as 
Qwen2.5-Math-7B-Instruct and fine-tuned variants including Qwen2.5-Math-7B-Base-SFT 
and Qwen2.5-7B-RAFT-Zero. These results demonstrate that BPO achieves both effective 
alignment and strong generalization, confirming its robust performance across different 
LLM architectures and model scales.

\begin{table}[!h]
    \centering
    \caption{Performance comparison across different model architectures and scales,
    it shows that BPO consistently outperforms DPO across all configurations and datasets.}
    \resizebox{1.0\textwidth}{!}{
    \begin{tabular}{lccccccc}
    \toprule
    \multirow{2}{*}{\textbf{Base Model}} &\multirow{2}{*}{\textbf{Method}} &  \multirow{2}{*}{\textbf{AIME2024}} & \multirow{2}{*}{\textbf{MATH500}} & \multirow{2}{*}{\textbf{AMC2023}} & \multirow{2}{*}{\begin{tabular}[c]{@{}c@{}} \bf Minerva \\ \bf Math \end{tabular}} & \multirow{2}{*}{\begin{tabular}[c]{@{}c@{}} \bf Olympiad \\ \bf Bench \end{tabular}} & \multirow{2}{*}{\textbf{Avg.}} \\
    & \\
    \midrule
    \multirow{2}{*}{Llama-3.1-8B-Instruct} & DPO & 3.3 & 44.6 & 12.5 & 22.1 & 11.6 & 18.8\\
    & BPO & {\bf 10.0} & {\bf 50.6} & {\bf 40.0} & {\bf 27.2} & {\bf 16.7} & {\bf 28.9} \\
    \midrule
    \multirow{2}{*}{Qwen2.5-Math-1.5B-Base} & DPO & 3.3 & 58.8 & 27.5 & {\bf 27.6} & 23.6 & 28.2 \\
     & BPO & {\bf 16.7} & {\bf 64.8} & {\bf 52.5} & 26.8 & {\bf 30.5} & {\bf 38.3} \\
    \midrule
    \multirow{2}{*}{Qwen2.5-Math-7B-Base} & DPO & 6.7 & 71.2 & 55.0 &  {\bf 39.3} & 32.9 & 41.0 \\
    & BPO & {\bf 30.0} & {\bf 75.8} & {\bf 60.0} & 31.2 & {\bf 36.3} & {\bf 46.7} \\
   \midrule
    \multirow{2}{*}{Qwen2.5-Math-7B-Instruct} & DPO & 10.0 & 77.0 & 60.0 & 28.7 & 38.1 & 42.8\\
    & BPO & {\bf 20.0} & {\bf 82.4} & {\bf 60.0} & {\bf 40.8} & {\bf 40.6} & {\bf 48.8} \\
    \bottomrule
    \end{tabular}
    }
    \label{tab:backbone}
    \end{table}

\begin{figure}[!h]
    \centering
    \includegraphics[width=1.0\textwidth]{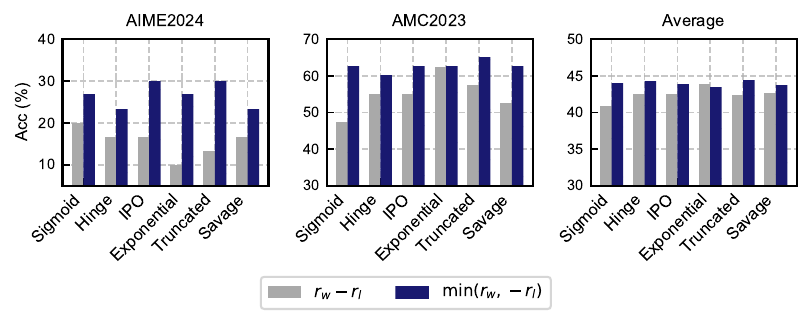}
    \vskip -0.6em
    \caption{
        Relative Reward Margin vs. Balanced Reward Margin under different preference 
        optimization functions. Balanced Reward Margin shows significant advantages 
        on AIME2024 and AMC2023.
        }
    \label{fig:loss_type}
\end{figure}

\paragraph{\textbf{Balanced reward margin is applicable to various preference optimization functions.}} \label{thm:various_loss} 
We investigate the applicability of the balanced reward margin across various preference 
optimization functions introduced in \cite{TangGZCMRRVPP24}.
As shown in Table~\ref{tab:loss_type}, the balanced reward 
margin consistently outperforms the relative margin across all tested optimization 
objectives. Specifically, under the logistic log loss used in DPO, the balanced margin 
improves performance from 41.0\% to 44.5\% (+3.5\%). Similarly, when applied with the hinge 
loss employed in SLiC, it achieves an improvement of +5.0\%, reaching 46.7\%. The balanced 
formulation also yields consistent gains of +1.9\% and +1.0\% under less commonly used losses 
such as truncated quadratic loss and Savage loss, respectively. Notably, as shown 
in Figure~\ref{fig:loss_type}, the balanced reward margin demonstrates significant 
advantages on challenging datasets like AIME2024 and AMC2023. These results indicate 
that the balanced reward margin provides more effective alignment with human preferences 
compared to the standard relative margin, highlighting its robustness and broad 
applicability across diverse preference learning settings.

\begin{table}[!h]
    \centering
    \caption{
    Performance comparison between the relative reward margin $x_1 - x_2$ and the 
    balanced reward margin $\min(x_1, -x_2)$ under different loss functions. The proposed 
    balanced reward margin shows consistent gains across various preference 
    optimization objectives. Gap adaptor is set to 0.3 in this experiment.
    }
    \resizebox{1.0\textwidth}{!}{
    \begin{tabular}{lccccc}
    \toprule
    \textbf{Loss Type}  &  \textbf{Algorithm} &  $\boldsymbol{f(\beta \rho_{\theta})}$& $\boldsymbol{x_1-x_2}$ & $\boldsymbol{\min(x_1,-x_2)}$& $\triangle$\\
    \midrule
    logistic log loss                   & DPO  & $\log(1+\exp(-\beta \rho_{\theta}))$ & 41.0 & 44.5 & \textcolor{blue}{+ 3.5} \\
    hinge loss                          & SLiC & $\max(0,1-\beta \rho_{\theta})$ & 41.7 & 46.7 & \textcolor{blue}{+ 5.0}  \\
    squared loss                        & IPO  & $(\beta \rho_{\theta}-1)^2$ & 43.1 & 43.9 & \textcolor{blue}{+ 0.8}  \\
    exponential loss                    & N/A   & $\exp(-\beta \rho_{\theta})$ & 43.5 & 43.9 & \textcolor{blue}{+ 0.4}  \\
    truncated quadratic loss            & N/A   & $(\max(0,1-\beta \rho_{\theta}))^2$ & 42.4 & 44.3 & \textcolor{blue}{+ 1.9}  \\
    savage loss                         & N/A   & $1/(1+\exp(\beta \rho_{\theta}))^2$ & 42.7 & 43.7 & \textcolor{blue}{+ 1.0}  \\
    \bottomrule
    \end{tabular}
    }
    \label{tab:loss_type}
    \end{table}

\begin{wrapfigure}{r}{0.5\textwidth}
    \vspace{-15pt}
    \centering
    \includegraphics[width=1.0\linewidth]{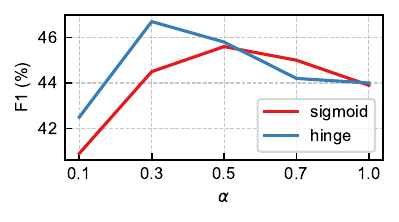}
    \vskip -1.0em
    \caption{
        Influence of the gap adaptor $\alpha$ on model performance. 
        Optimal performance is achieved when the gap adaptor is set to a moderate value.
    }
    \vspace{-9pt} 
    \label{fig:alpha}
\end{wrapfigure}

\paragraph{\textbf{Moderately relaxing the constraints on rejected responses can enhance model performance.}} \label{thm:gap_adaptor}
We analyzed the impact of the gap adaptor ($\alpha$) parameter on model performance. 
As shown in Figure 5, BPO performance initially increases and then decreases as 
$\alpha$ increases. For the logistic log loss , the best performance is achieved when 
$\alpha = 0.5$. In contrast, for the hinge loss, the optimal performance is observed 
at $\alpha = 0.3$. These findings suggest that it is not necessary to strictly balance 
rewards of chosen responses and rejected responses. Instead, moderately relaxing the constraints 
on rejected responses can enhance model performance.

\section{Related Work}
\textbf{Direct Preference Optimization.}  
Efficient preference optimization algorithms, such as DPO have emerged as a popular approach for 
aligning LLMs with human preferences. Compared to traditional methods like RLHF, DPO offers several 
advantages, including greater stability, strong performance, and improved computational efficiency.
Despite these benefits, recent studies have identified several challenges associated with DPO. For 
instance, its implicit reward modeling can lead to biased policies that favor out-of-distribution 
responses \cite{XuFGYLMW0024,abs-2404-14723}. Additionally, offline DPO has been found to be 
empirically less effective than online alignment methods \cite{IvisonW0WP0S0H24}, and aligned models 
may suffer from degrading performance after alignment \cite{LinL0DLZP00ZDPZ24,abs-2405-17931}. 
In response to these limitations, various enhanced versions of DPO have been proposed, including  
CPO \cite{XuSCTSDM024}, KTO \cite{EthayarajhXMJK24}, IPO \cite{AzarGPMRVC24},
SLiC \cite{0002ZJKSLL24}, Cal-DPO \cite{XiaoYZLH24} and DPOP \cite{abs-2402-13228}. 
However, none of these approaches fundamentally address the DCR problem, which limits the potential 
performance improvements of DPO.

\textbf{Bradley-Terry Model.} The Bradley-Terry (BT) model \cite{bradley1952rank} was originally 
proposed to convert pairwise comparisons into numerical scores. With the development of reinforcement 
learning from human feedback (RLHF) \cite{StiennonO0ZLVRA20,Ouyang0JAWMZASR22,abs-2204-05862}
the BT model has been widely used to optimize reward models, achieving significant success in improving 
the performance of LLMs across various tasks. Later, with the introduction of DPO, the BT model was 
further applied to model preferences directly for LLM preference tuning.
However, several studies have discussed the limitations and challenges of using the BT model within 
the RLHF framework from different perspectives \cite{AzarGPMRVC24,TangGZCMRRVPP24,abs-2305-10425}.
For instance, it has been noted that the Bradley-Terry (BT) model cannot capture non-transitive 
preferences, and maximizing the corresponding Elo score may not align with the true objective 
of preference optimization \cite{MunosVCARGTGMFM24}. Additionally, applying the BT model within 
DPO can result in problematic overfitting when the observed preferences are deterministic 
\cite{AzarGPMRVC24}. In our work, we revisit the foundational mechanisms of the BT model and 
propose improvements to address these limitations.

\section{Conclusion and Limitations}\label{sec:conclusion}
In this work, we identify a critical issue in direct preference optimization methods: 
the Degraded Chosen Responses (DCR) problem. To address this limitation, we propose BPO, a simple yet effective framework 
that explicitly preserves the absolute likelihood of chosen responses while still maximizing 
the reward gap between chosen and rejected responses. BPO can be seamlessly integrated into 
existing preference optimization algorithms with 
minimal modifications, offering both theoretical guarantees and practical benefits. 
We hope it inspires further 
research into more balanced and effective preference learning objectives.

A limitation of BPO is that it is currently restricted to offline methods and does not 
incorporate on-policy learning, where the policy can interact with the reward model during training. 
It would be interesting to explore how the balanced reward margin approach used in BPO performs 
in an on-policy learning scenario. We consider this an exciting direction for future research.

\bibliographystyle{plain}
\bibliography{references} 

\newpage
\appendix

\section{Derivations and Proofs}
\label{sec:derivations}

\subsection{Gradient of BPO Loss}
\label{sec:gradient}

We consider the following BPO loss function:

\begin{equation}
\mathcal{L}(\theta) = -\mathbb{E}_{(\mathbf{x}, \mathbf{y}_w, \mathbf{y}_l) \sim D} \left[ f\left( \min\left( \beta \log \frac{\pi_\theta(\mathbf{y}_w \mid \mathbf{x})}{\pi_{\text{ref}}(\mathbf{y}_w \mid \mathbf{x})},\ -\alpha \beta \log \frac{\pi_\theta(\mathbf{y}_l \mid \mathbf{x})}{\pi_{\text{ref}}(\mathbf{y}_l \mid \mathbf{x})} \right) \right) \right].
\end{equation}

The loss function can be simplified as follows:

\begin{equation}
\mathcal{L}(\theta) = -\mathbb{E}_{(\mathbf{x}, \mathbf{y}_w, \mathbf{y}_l)} \left[ f\left( \min(\beta r_w,\ -\alpha \beta r_l) \right) \right].
\end{equation}

Where,

\begin{equation}
r_w = \log \frac{\pi_\theta(\mathbf{y}_w \mid \mathbf{x})}{\pi_{\text{ref}}(\mathbf{y}_w \mid \mathbf{x})}, \quad
r_l = \log \frac{\pi_\theta(\mathbf{y}_l \mid \mathbf{x})}{\pi_{\text{ref}}(\mathbf{y}_l \mid \mathbf{x})}.
\end{equation}

Let: $z = \min(\beta r_w,\ -\alpha \beta r_l)$,
we compute the gradient:
\begin{equation}
\nabla_\theta \mathcal{L}(\theta) = -\nabla_\theta \mathbb{E}[f(z)] = -\mathbb{E}[\nabla_\theta f(z)].
\end{equation}

By the chain rule:
$\nabla_\theta f(z) = f'(z) \cdot \nabla_\theta z$, thus:
\begin{equation}
\nabla_\theta \mathcal{L}(\theta) = -\mathbb{E}[f'(z) \cdot \nabla_\theta z].
\end{equation}

To proceed, we analyze $\nabla_\theta z$, which depends on which argument achieves the minimum in the definition of $z$.

Case 1: $\beta r_w < -\alpha \beta r_l$, then $z = \beta r_w$, so:
\begin{equation}
\nabla_\theta f(z) = f'(\beta r_w) \cdot \beta \cdot \nabla_\theta \log \pi_\theta(\mathbf{y}_w \mid \mathbf{x}).
\end{equation}

Case 2: $\beta r_w \geq -\alpha \beta r_l$, then $z = -\alpha \beta r_l$, so:
\begin{equation}
\nabla_\theta f(z) = f'(-\alpha \beta r_l) \cdot (-\alpha \beta) \cdot \nabla_\theta \log \pi_\theta(\mathbf{y}_l \mid \mathbf{x}).
\end{equation}

Combining both cases, the gradient of BPO loss function is:

\begin{equation}
\boxed{
\nabla_\theta \mathcal{L}(\theta) =
-\mathbb{E}_{(\mathbf{x}, \mathbf{y}_w, \mathbf{y}_l)} 
\begin{cases}
f'(\beta r_w) \cdot \beta \cdot \nabla_\theta \log \pi_\theta(\mathbf{y}_w \mid \mathbf{x}), & \text{if } \beta r_w < -\alpha \beta r_l, \\
f'(-\alpha \beta r_l) \cdot (-\alpha \beta) \cdot \nabla_\theta \log \pi_\theta(\mathbf{y}_l \mid \mathbf{x}), & \text{otherwise}.
\end{cases}
}
\end{equation}

\subsection{Proofs of Theorem 1}
\label{sec:proof_theorem1}

BPO maximizes $ \rho^b_\theta $, where $ \gamma $ denotes the margin achieved during optimization such that:
$\rho^b_\theta \geq \gamma.$

Recall that:
\begin{equation}
\rho^b_\theta = \min(r_w, -\alpha r_l). 
\end{equation}

For this minimum to be at least $ \gamma $, both components must individually satisfy:
\begin{equation}
r_w \geq \gamma \quad \text{and} \quad -\alpha r_l \geq \gamma.
\end{equation}

Focusing on the first inequality, $ r_w \geq \gamma $, and expanding $ r_w $ gives:
\begin{equation}
\beta \log \frac{\pi_\theta(\mathbf{y}_w | \mathbf{x})}{\pi_{\text{ref}}(\mathbf{y}_w | \mathbf{x})} \geq \gamma.
\end{equation}

Dividing both sides by $ \beta $ (noting that $ \beta > 0 $):
\begin{equation}
\log \frac{\pi_\theta(\mathbf{y}_w | \mathbf{x})}{\pi_{\text{ref}}(\mathbf{y}_w | \mathbf{x})} \geq \frac{\gamma}{\beta}.
\end{equation}

Applying the exponential function to both sides to eliminate the logarithm:
\begin{equation}
\frac{\pi_\theta(\mathbf{y}_w | \mathbf{x})}{\pi_{\text{ref}}(\mathbf{y}_w | \mathbf{x})} \geq \exp\left(\frac{\gamma}{\beta}\right).
\end{equation}

Finally, rearranging yields the desired result:
\begin{equation}
\boxed{
\pi_\theta(\mathbf{y}_w | \mathbf{x}) \geq \exp\left(\frac{\gamma}{\beta}\right) \pi_{\text{ref}}(\mathbf{y}_w | \mathbf{x}).
}
\end{equation}


\section{Prompt}
\label{sec:prompt}

The prompt used for the policy generator is shown below.

\begin{tcolorbox}[title=Prompt for policy generator, label={tab:rationale_prompt}, breakable, width=\textwidth,
fonttitle=\bfseries
]
\textbf{[System]:} \\
Please reason step by step, and put your final answer within boxed\{\}. \\

\textbf{[User]:} \\
The following is the math problem: \\

[Math Problem] \\

\{problem\} \\

Let's think step by step and output the final answer within boxed\{\}. \\
\end{tcolorbox}


\section{Additional Results}
\begin{figure}[H]
    \centering
    \includegraphics[width=1.0\textwidth]{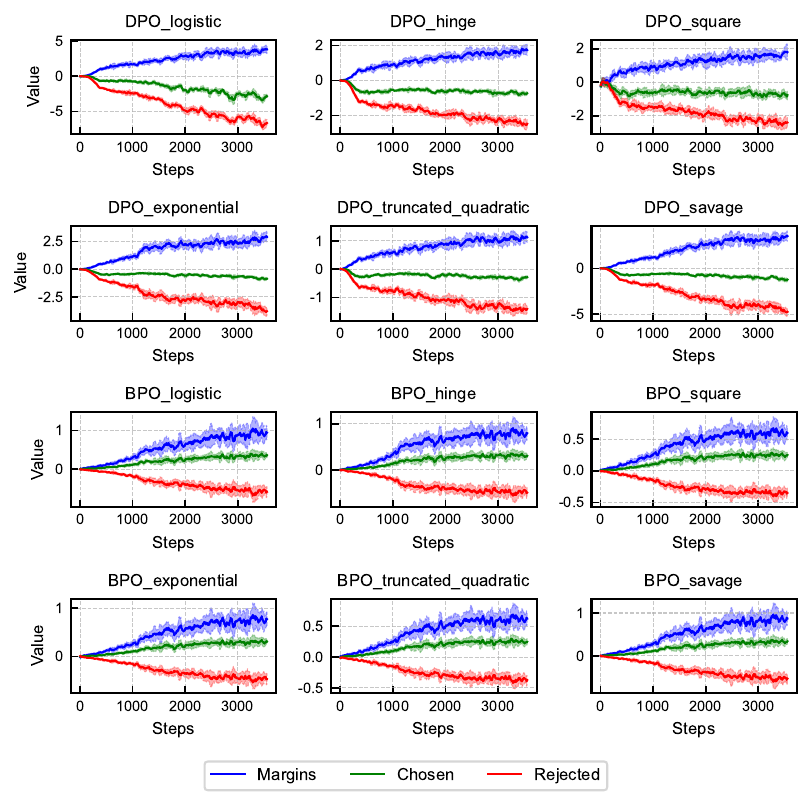}
    \caption{
    The reward dynamics of both chosen and rejected responses, along with the evolution of their margins. These results demonstrate that BPO consistently maintains positive 
    $r_w$ values, which continue to increase during training. Note that the gap adaptor is set to 1.0 for these comparisons.
    }
    \label{fig:interaction_appendix}
\end{figure}

We present the reward dynamics of chosen responses ($r_w$) and rejected responses ($r_l$), 
as well as the evolution of their margins ($r_w - r_l$), for both DPO and BPO using different 
loss functions during training. As shown in Figure \ref{fig:interaction_appendix}, despite 
various efforts to improve DPO, none of the existing variants effectively address the DCR problem. 
In these methods, $r_w$ consistently declines throughout training, which contradicts the 
intended training objective.
In contrast, our proposed method, BPO, successfully mitigates the DCR issue. When applied 
with different preference optimization functions, BPO ensures that all $r_w$ values 
remain positive and continue to increase during training. This demonstrates the clear 
advantage of using a balanced reward margin over the traditional relative reward margin 
in preference optimization.

\section{Societal Impacts}
\label{app:societal-impacts}
We propose Balanced Preference Optimization (BPO) to align large language models (LLMs) with 
human preferences, thereby fostering more reliable and socially aligned AI systems. By enhancing 
training stability and efficiency, BPO addresses key limitations in direct preference optimization 
methods. Given the widespread deployment of LLMs in various real-world applications that interact 
closely with humans, it is essential to ensure that these models' responses to prompts are aligned 
with human and societal values. This alignment can have significant positive social impacts.


\section{Case Study}
This section presents comparative examples between BPO and the baseline DPO algorithm, 
along with its variant IPO, the results are shown below in 
Tables~\ref{tab:dialogue-ex-1} and~\ref{tab:dialogue-ex-2}.

\begin{table}[!h]
    \centering
    \ttfamily
    \caption{Examples comparing BPO with the baseline DPO show that BPO arrives at the correct solution more concisely.}
    \begin{tabular}{lp{11.5cm}}
        \toprule
        \bf{Prompt} & Human: Find the roots of $(x - 3)^3 + (x -7)^3 = (2x - 10)^3$. Assistant:\\
        \midrule
        \bf{DPO} & First, let's simplify the equation. We can make a substitution to make the equation easier to handle. Let $ y = x - 5 $. This substitution is chosen because 5 is the midpoint.
            When $ x = y + 5 $, the equation becomes:
            $(y + 5 - 3)^3 + (y + 5 - 7)^3 = (2(y + 5) - 10)^3$
            $(y + 2)^3 + (y - 2)^3 = (2y)^3$
            This simplifies.
            Let
            $(y+2)^3 + (y-2)^3 = (2y)^3$
            Expanding the cubes:
            $(y+2)^3 = y^3 + 12y^2 + 8y + 8$
            $(y-2)^3 = y^3 - 12y^2 - 8y + 8$
            So,
            $(y+2)^3 + (y-2)^3 = y^3 + 12y^2 + 8y + 8 + y^3 - 12y^2 - 8y + 8 = 8y^3$
            We can simplify this by combining the terms:
            $y^3 + 12y^2 + 8y + 8 + y^3 - 12y^2 - 8y + 8 = 8y^3$
            $2y^3 + 16 = 8y^3$
            Rearranging terms, we
            $ 8y^3 - 2y^3 = 16 \implies 6y^3 = 16$
            Simplify further:
            $y^3 = \frac{16}{6} = \frac{8}{3} \text{ or there is a simpler transformation we, look at the structure.}$
            Instead, let's look at the equation in a different way. Notice that the equation $ (y+2)^3 + (y-2)^3 = (2y)^3 $ can.
            Notice that if we let $a = y+2$, $b = y-2$, and $c = 2y$, we.
            One obvious solution to
            $a^3 + b^3 = c^3$
            we recall that equations like $ (y+2)^3 + (y-2)^3 = (2y)^3 $ often resolves to simple substitutions or specific values. Not directly, we see that:
            If $ y = 0 $, then $(2)^3 + (-2)^3 = 0^3$, which is $8 - 8 = 0$, which is true.
            Another way to see
            Let $ y = x - 5 $. If $ y = 0 $, then $ x = 5 $. We can substitute $ x - 5 = 0 $
            If $ x = 5 $: $x - 3 = 2$, $x - 7 = -2$, $2x - 10 = 0$
            Clearly, $ (2)^3 + (-2)^3 = 0^3 $, which is $8 - 8 = 0$, and it
            So,
            $x = 5$ is a root. We can use.
            Thus,
            Let's box,
            $\boxed{5}$ is the solution. \hspace{1em}\scalebox{2}{\textcolor{red}{\faTimes}}
        \\
        \midrule
        \bf{BPO} & To solve the equation $(x - 3)^3 + (x - 7)^3 = (2x - 10)^3$, we can use a substitution to simplify the equation. Let's set $y = x - 5$. This substitution is chosen because it centers the terms around a common midpoint, which can simplify the algebra.
        With this substitution, we have:
        $x - 3 = y + 2,$
        $x - 7 = y - 2,$
        $2x - 10 = 2y.$
        Substituting these into the original equation, we get:
        $(y + 2)^3 + (y - 2)^3 = (2y)^3.$
        Expanding both sides, we have:
        $y^3 + 6y^2 + 12y + 8 + y^3 - 6y^2 + 12y - 8 = 8y^3,$
        $2y^3 + 24y = 8y^3,$
        $24y = 6y^3,$
        $4y = y^3,$
        $y^3 - 4y = 0,$
        $y(y^2 - 4) = 0,$
        $y(y - 2)(y + 2) = 0.$
        So, the solutions for $y$ are $y = 0$, $y = 2$, and $y = -2$. Converting back to $x$, we have:
        $x = 5 + 0 = 5,$
        $x = 5 + 2 = 7,$
        $x = 5 - 2 = 3.$ 
        So, the final answer is:
        $\boxed{3, 5, 7}$  \hspace{1em}\scalebox{2}{\textcolor{green}{\faCheck}} \\
        \bottomrule
    \end{tabular}
    \label{tab:dialogue-ex-1}
\end{table}

\begin{table}[!h]
    \centering
    \ttfamily
    \caption{Example comparing BPO with the the DPO variant IPO.}
    \begin{tabular}{lp{11.5cm}}
        \toprule
        \bf{Prompt} & Human: For some real numbers $a$ and $b$, the equation $8x^3 + 4ax^2 + 2bx + a = 0$
        has three distinct positive roots. If the sum of the base-2 logarithms of the roots is 5, what is the value of $a$?
        Assistant:\\
        \midrule
        \bf{IPO} & Given the equation $8x^3 + 4ax^2 + 2bx + a = 0$ with roots $r_1, r_2, r_3$, 
        we are also given that the roots are positive and distinct. 
        By the properties of logarithms and the problem statement, we, know that: 
        $
        \log_2(r_1) + \log_2(r_2) + \log_2(r_3) = 5
        $ 
        Using the logarithm property that $\log_2(r_1 r_2 r_3) = \log_2(r_1) + \log_2(r_2) + \log_2(r_3)$, we. 
        can deduce that: 
        $
        \log_2(r_1 r_2 r_3) = 5 \implies r_1 r_2 r_3 = 2^5 = 32,
        $ 
        ,which means the product of the roots $r_1 r_2 r _3 = 32$. 
        By Vieta's formulas for the polynomial $8x^3 + 4ax^2 + 2bx + a = 0$, the product of the roots 
        for a polynomial $Ax^3 + Bx^2 + Cx + D = 0$ is given by $-\frac{D}{A}$. Here, the leading coefficient 
        $A = 8$ and the constant term is $a$, so: 
        $
        r_1 r_2 r_3 = -\frac{a}{-8} = \frac{a}{8} \implies \frac{a}{8} = 32 \implies a = 8 \cdot 32 = 256.
        $ 
        Thus, the value of $a$ is $256$. The final answer is: 
        $
        \boxed{256}
        $  \hspace{1em}\scalebox{2}{\textcolor{red}{\faTimes}}
        \\
        \midrule
        \bf{BPO} & 
        To solve the problem, we need to use the properties of polynomials and logarithms. Here's the step-by-step reasoning:
        1. Sum of the Base-2 Logarithms of the Roots:
        If the roots of the polynomial $8x^3 + 4a x^2 + 2b x + a = 0$ are $r_1, r_2, r_3$, then the sum of the base-2 logarithms of the roots is given by:
        $\log_2(r_1) + \log_2(r_2) + \log_2(r_3) = 5$
        Using the properties of logarithms, this can be rewritten as:
        $\log_2(r_1 r_2 r_3) = 5$
        This implies:
        $r_1 r_2 r_3 = 2^5 = 32$
        2. Product of the Roots: 
        By Vieta's formulas for the polynomial $8x^3 + 4a x^2 + 2b x + a = 0$, the product of the roots $r_1, r_2, r_3$ is given by:
        $r_1 r_2 r_3 = -\frac{a}{8}$
        From the previous step, we know $r_1 r_2 r_3 = 32$, so:
        $-\frac{a}{8} = 32$
        Solving for $a$:
        $a = -256$
        The value of $a$ is $\boxed{-256}$.  \hspace{1em}\scalebox{2}{\textcolor{green}{\faCheck}}
        \\
        \bottomrule
    \end{tabular}
    \label{tab:dialogue-ex-2}
\end{table}

\clearpage
\newpage
\section*{NeurIPS Paper Checklist}

\begin{enumerate}

\item {\bf Claims}
    \item[] Question: Do the main claims made in the abstract and introduction accurately reflect the paper's contributions and scope?
    \item[] Answer: \answerYes{} 
    \item[] Justification: Our main claim matches our theoretical and experimental results in Section~\ref{sec:theoretical_analysis} and Section~\ref{sec:experiment}.
    \item[] Guidelines:
    \begin{itemize}
        \item The answer NA means that the abstract and introduction do not include the claims made in the paper.
        \item The abstract and/or introduction should clearly state the claims made, including the contributions made in the paper and important assumptions and limitations. A No or NA answer to this question will not be perceived well by the reviewers. 
        \item The claims made should match theoretical and experimental results, and reflect how much the results can be expected to generalize to other settings. 
        \item It is fine to include aspirational goals as motivation as long as it is clear that these goals are not attained by the paper. 
    \end{itemize}

\item {\bf Limitations}
    \item[] Question: Does the paper discuss the limitations of the work performed by the authors?
    \item[] Answer: \answerYes{} 
    \item[] Justification: Please see Section~\ref{sec:conclusion} for the discussion of limitations.
    \item[] Guidelines:
    \begin{itemize}
        \item The answer NA means that the paper has no limitation while the answer No means that the paper has limitations, but those are not discussed in the paper. 
        \item The authors are encouraged to create a separate "Limitations" section in their paper.
        \item The paper should point out any strong assumptions and how robust the results are to violations of these assumptions (e.g., independence assumptions, noiseless settings, model well-specification, asymptotic approximations only holding locally). The authors should reflect on how these assumptions might be violated in practice and what the implications would be.
        \item The authors should reflect on the scope of the claims made, e.g., if the approach was only tested on a few datasets or with a few runs. In general, empirical results often depend on implicit assumptions, which should be articulated.
        \item The authors should reflect on the factors that influence the performance of the approach. For example, a facial recognition algorithm may perform poorly when image resolution is low or images are taken in low lighting. Or a speech-to-text system might not be used reliably to provide closed captions for online lectures because it fails to handle technical jargon.
        \item The authors should discuss the computational efficiency of the proposed algorithms and how they scale with dataset size.
        \item If applicable, the authors should discuss possible limitations of their approach to address problems of privacy and fairness.
        \item While the authors might fear that complete honesty about limitations might be used by reviewers as grounds for rejection, a worse outcome might be that reviewers discover limitations that aren't acknowledged in the paper. The authors should use their best judgment and recognize that individual actions in favor of transparency play an important role in developing norms that preserve the integrity of the community. Reviewers will be specifically instructed to not penalize honesty concerning limitations.
    \end{itemize}

\item {\bf Theory assumptions and proofs}
    \item[] Question: For each theoretical result, does the paper provide the full set of assumptions and a complete (and correct) proof?
    \item[] Answer: \answerYes{} 
    \item[] Justification: Please refer to Section~\ref{sec:theoretical_analysis} and Appendix~\ref{sec:derivations} for our assumptions and a complete (and correct) proof.
    \item[] Guidelines:
    \begin{itemize}
        \item The answer NA means that the paper does not include theoretical results. 
        \item All the theorems, formulas, and proofs in the paper should be numbered and cross-referenced.
        \item All assumptions should be clearly stated or referenced in the statement of any theorems.
        \item The proofs can either appear in the main paper or the supplemental material, but if they appear in the supplemental material, the authors are encouraged to provide a short proof sketch to provide intuition. 
        \item Inversely, any informal proof provided in the core of the paper should be complemented by formal proofs provided in appendix or supplemental material.
        \item Theorems and Lemmas that the proof relies upon should be properly referenced. 
    \end{itemize}

    \item {\bf Experimental result reproducibility}
    \item[] Question: Does the paper fully disclose all the information needed to reproduce the main experimental results of the paper to the extent that it affects the main claims and/or conclusions of the paper (regardless of whether the code and data are provided or not)?
    \item[] Answer: \answerYes{} 
    \item[] Justification: We provide information needed to reproduce the main experimental result and the link which contains the code and dataset to reproduce our results.
    \item[] Guidelines:
    \begin{itemize}
        \item The answer NA means that the paper does not include experiments.
        \item If the paper includes experiments, a No answer to this question will not be perceived well by the reviewers: Making the paper reproducible is important, regardless of whether the code and data are provided or not.
        \item If the contribution is a dataset and/or model, the authors should describe the steps taken to make their results reproducible or verifiable. 
        \item Depending on the contribution, reproducibility can be accomplished in various ways. For example, if the contribution is a novel architecture, describing the architecture fully might suffice, or if the contribution is a specific model and empirical evaluation, it may be necessary to either make it possible for others to replicate the model with the same dataset, or provide access to the model. In general. releasing code and data is often one good way to accomplish this, but reproducibility can also be provided via detailed instructions for how to replicate the results, access to a hosted model (e.g., in the case of a large language model), releasing of a model checkpoint, or other means that are appropriate to the research performed.
        \item While NeurIPS does not require releasing code, the conference does require all submissions to provide some reasonable avenue for reproducibility, which may depend on the nature of the contribution. For example
        \begin{enumerate}
            \item If the contribution is primarily a new algorithm, the paper should make it clear how to reproduce that algorithm.
            \item If the contribution is primarily a new model architecture, the paper should describe the architecture clearly and fully.
            \item If the contribution is a new model (e.g., a large language model), then there should either be a way to access this model for reproducing the results or a way to reproduce the model (e.g., with an open-source dataset or instructions for how to construct the dataset).
            \item We recognize that reproducibility may be tricky in some cases, in which case authors are welcome to describe the particular way they provide for reproducibility. In the case of closed-source models, it may be that access to the model is limited in some way (e.g., to registered users), but it should be possible for other researchers to have some path to reproducing or verifying the results.
        \end{enumerate}
    \end{itemize}

\item {\bf Open access to data and code}
    \item[] Question: Does the paper provide open access to the data and code, with sufficient instructions to faithfully reproduce the main experimental results, as described in supplemental material?
    \item[] Answer: \answerYes{}  
    \item[] Justification: The code of BPO is available
    \item[] Guidelines:
    \begin{itemize}
        \item The answer NA means that paper does not include experiments requiring code.
        \item Please see the NeurIPS code and data submission guidelines (\url{https://nips.cc/public/guides/CodeSubmissionPolicy}) for more details.
        \item While we encourage the release of code and data, we understand that this might not be possible, so “No” is an acceptable answer. Papers cannot be rejected simply for not including code, unless this is central to the contribution (e.g., for a new open-source benchmark).
        \item The instructions should contain the exact command and environment needed to run to reproduce the results. See the NeurIPS code and data submission guidelines (\url{https://nips.cc/public/guides/CodeSubmissionPolicy}) for more details.
        \item The authors should provide instructions on data access and preparation, including how to access the raw data, preprocessed data, intermediate data, and generated data, etc.
        \item The authors should provide scripts to reproduce all experimental results for the new proposed method and baselines. If only a subset of experiments are reproducible, they should state which ones are omitted from the script and why.
        \item At submission time, to preserve anonymity, the authors should release anonymized versions (if applicable).
        \item Providing as much information as possible in supplemental material (appended to the paper) is recommended, but including URLs to data and code is permitted.
    \end{itemize}

\item {\bf Experimental setting/details}
    \item[] Question: Does the paper specify all the training and test details (e.g., data splits, hyperparameters, how they were chosen, type of optimizer, etc.) necessary to understand the results?
    \item[] Answer: \answerYes{} 
    \item[] Justification: We provide all the training and test details in Section~\ref{sec:experiment}.
    \item[] Guidelines:
    \begin{itemize}
        \item The answer NA means that the paper does not include experiments.
        \item The experimental setting should be presented in the core of the paper to a level of detail that is necessary to appreciate the results and make sense of them.
        \item The full details can be provided either with the code, in appendix, or as supplemental material.
    \end{itemize}

\item {\bf Experiment statistical significance}
    \item[] Question: Does the paper report error bars suitably and correctly defined or other appropriate information about the statistical significance of the experiments?
    \item[] Answer: \answerYes{} 
    \item[] Justification: We provide the experiment results that support the main claims of the paper.
    \item[] Guidelines:
    \begin{itemize}
        \item The answer NA means that the paper does not include experiments.
        \item The authors should answer "Yes" if the results are accompanied by error bars, confidence intervals, or statistical significance tests, at least for the experiments that support the main claims of the paper.
        \item The factors of variability that the error bars are capturing should be clearly stated (for example, train/test split, initialization, random drawing of some parameter, or overall run with given experimental conditions).
        \item The method for calculating the error bars should be explained (closed form formula, call to a library function, bootstrap, etc.)
        \item The assumptions made should be given (e.g., Normally distributed errors).
        \item It should be clear whether the error bar is the standard deviation or the standard error of the mean.
        \item It is OK to report 1-sigma error bars, but one should state it. The authors should preferably report a 2-sigma error bar than state that they have a 96\% CI, if the hypothesis of Normality of errors is not verified.
        \item For asymmetric distributions, the authors should be careful not to show in tables or figures symmetric error bars that would yield results that are out of range (e.g. negative error rates).
        \item If error bars are reported in tables or plots, The authors should explain in the text how they were calculated and reference the corresponding figures or tables in the text.
    \end{itemize}

\item {\bf Experiments compute resources}
    \item[] Question: For each experiment, does the paper provide sufficient information on the computer resources (type of compute workers, memory, time of execution) needed to reproduce the experiments?
    \item[] Answer: \answerYes{} 
    \item[] Justification: We provide the computer resources in Section~\ref{sec:experiment}.
    \item[] Guidelines:
    \begin{itemize}
        \item The answer NA means that the paper does not include experiments.
        \item The paper should indicate the type of compute workers CPU or GPU, internal cluster, or cloud provider, including relevant memory and storage.
        \item The paper should provide the amount of compute required for each of the individual experimental runs as well as estimate the total compute. 
        \item The paper should disclose whether the full research project required more compute than the experiments reported in the paper (e.g., preliminary or failed experiments that didn't make it into the paper). 
    \end{itemize}
    
\item {\bf Code of ethics}
    \item[] Question: Does the research conducted in the paper conform, in every respect, with the NeurIPS Code of Ethics \url{https://neurips.cc/public/EthicsGuidelines}?
    \item[] Answer: \answerYes{} 
    \item[] Justification: We make sure to preserve anonymity and conform NeurIPS Code of Ethics.
    \item[] Guidelines:
    \begin{itemize}
        \item The answer NA means that the authors have not reviewed the NeurIPS Code of Ethics.
        \item If the authors answer No, they should explain the special circumstances that require a deviation from the Code of Ethics.
        \item The authors should make sure to preserve anonymity (e.g., if there is a special consideration due to laws or regulations in their jurisdiction).
    \end{itemize}

\item {\bf Broader impacts}
    \item[] Question: Does the paper discuss both potential positive societal impacts and negative societal impacts of the work performed?
    \item[] Answer: \answerYes{} 
    \item[] Justification:Please see Appendix~\ref{app:societal-impacts} for broader impacts.
    \item[] Guidelines:
    \begin{itemize}
        \item The answer NA means that there is no societal impact of the work performed.
        \item If the authors answer NA or No, they should explain why their work has no societal impact or why the paper does not address societal impact.
        \item Examples of negative societal impacts include potential malicious or unintended uses (e.g., disinformation, generating fake profiles, surveillance), fairness considerations (e.g., deployment of technologies that could make decisions that unfairly impact specific groups), privacy considerations, and security considerations.
        \item The conference expects that many papers will be foundational research and not tied to particular applications, let alone deployments. However, if there is a direct path to any negative applications, the authors should point it out. For example, it is legitimate to point out that an improvement in the quality of generative models could be used to generate deepfakes for disinformation. On the other hand, it is not needed to point out that a generic algorithm for optimizing neural networks could enable people to train models that generate Deepfakes faster.
        \item The authors should consider possible harms that could arise when the technology is being used as intended and functioning correctly, harms that could arise when the technology is being used as intended but gives incorrect results, and harms following from (intentional or unintentional) misuse of the technology.
        \item If there are negative societal impacts, the authors could also discuss possible mitigation strategies (e.g., gated release of models, providing defenses in addition to attacks, mechanisms for monitoring misuse, mechanisms to monitor how a system learns from feedback over time, improving the efficiency and accessibility of ML).
    \end{itemize}
    
\item {\bf Safeguards}
    \item[] Question: Does the paper describe safeguards that have been put in place for responsible release of data or models that have a high risk for misuse (e.g., pretrained language models, image generators, or scraped datasets)?
    \item[] Answer: \answerNA{} 
    \item[] Justification: The paper poses no such risks.
    \item[] Guidelines:
    \begin{itemize}
        \item The answer NA means that the paper poses no such risks.
        \item Released models that have a high risk for misuse or dual-use should be released with necessary safeguards to allow for controlled use of the model, for example by requiring that users adhere to usage guidelines or restrictions to access the model or implementing safety filters. 
        \item Datasets that have been scraped from the Internet could pose safety risks. The authors should describe how they avoided releasing unsafe images.
        \item We recognize that providing effective safeguards is challenging, and many papers do not require this, but we encourage authors to take this into account and make a best faith effort.
    \end{itemize}

\item {\bf Licenses for existing assets}
    \item[] Question: Are the creators or original owners of assets (e.g., code, data, models), used in the paper, properly credited and are the license and terms of use explicitly mentioned and properly respected?
    \item[] Answer: \answerNA{} 
    \item[] Justification: The paper does not use existing assets.
    \item[] Guidelines:
    \begin{itemize}
        \item The answer NA means that the paper does not use existing assets.
        \item The authors should cite the original paper that produced the code package or dataset.
        \item The authors should state which version of the asset is used and, if possible, include a URL.
        \item The name of the license (e.g., CC-BY 4.0) should be included for each asset.
        \item For scraped data from a particular source (e.g., website), the copyright and terms of service of that source should be provided.
        \item If assets are released, the license, copyright information, and terms of use in the package should be provided. For popular datasets, \url{paperswithcode.com/datasets} has curated licenses for some datasets. Their licensing guide can help determine the license of a dataset.
        \item For existing datasets that are re-packaged, both the original license and the license of the derived asset (if it has changed) should be provided.
        \item If this information is not available online, the authors are encouraged to reach out to the asset's creators.
    \end{itemize}

\item {\bf New assets}
    \item[] Question: Are new assets introduced in the paper well documented and is the documentation provided alongside the assets?
    \item[] Answer: \answerNA{} 
    \item[] Justification: The paper does not release new assets.
    \item[] Guidelines:
    \begin{itemize}
        \item The answer NA means that the paper does not release new assets.
        \item Researchers should communicate the details of the dataset/code/model as part of their submissions via structured templates. This includes details about training, license, limitations, etc. 
        \item The paper should discuss whether and how consent was obtained from people whose asset is used.
        \item At submission time, remember to anonymize your assets (if applicable). You can either create an anonymized URL or include an anonymized zip file.
    \end{itemize}

\item {\bf Crowdsourcing and research with human subjects}
    \item[] Question: For crowdsourcing experiments and research with human subjects, does the paper include the full text of instructions given to participants and screenshots, if applicable, as well as details about compensation (if any)? 
    \item[] Answer: \answerNA{} 
    \item[] Justification: The paper does not involve crowdsourcing nor research with human subjects.
    \item[] Guidelines:
    \begin{itemize}
        \item The answer NA means that the paper does not involve crowdsourcing nor research with human subjects.
        \item Including this information in the supplemental material is fine, but if the main contribution of the paper involves human subjects, then as much detail as possible should be included in the main paper. 
        \item According to the NeurIPS Code of Ethics, workers involved in data collection, curation, or other labor should be paid at least the minimum wage in the country of the data collector. 
    \end{itemize}

\item {\bf Institutional review board (IRB) approvals or equivalent for research with human subjects}
    \item[] Question: Does the paper describe potential risks incurred by study participants, whether such risks were disclosed to the subjects, and whether Institutional Review Board (IRB) approvals (or an equivalent approval/review based on the requirements of your country or institution) were obtained?
    \item[] Answer: \answerNA{} 
    \item[] Justification: The paper does not involve crowdsourcing nor research with human subjects.
    \item[] Guidelines:
    \begin{itemize}
        \item The answer NA means that the paper does not involve crowdsourcing nor research with human subjects.
        \item Depending on the country in which research is conducted, IRB approval (or equivalent) may be required for any human subjects research. If you obtained IRB approval, you should clearly state this in the paper. 
        \item We recognize that the procedures for this may vary significantly between institutions and locations, and we expect authors to adhere to the NeurIPS Code of Ethics and the guidelines for their institution. 
        \item For initial submissions, do not include any information that would break anonymity (if applicable), such as the institution conducting the review.
    \end{itemize}

\item {\bf Declaration of LLM usage}
    \item[] Question: Does the paper describe the usage of LLMs if it is an important, original, or non-standard component of the core methods in this research? Note that if the LLM is used only for writing, editing, or formatting purposes and does not impact the core methodology, scientific rigorousness, or originality of the research, declaration is not required.
    \item[] Answer: \answerNA{} 
    \item[] Justification: The core method development in this research does not involve LLMs as any important, original, or non-standard components.
    \item[] Guidelines:
    \begin{itemize}
        \item The answer NA means that the core method development in this research does not involve LLMs as any important, original, or non-standard components.
        \item Please refer to our LLM policy (\url{https://neurips.cc/Conferences/2025/LLM}) for what should or should not be described.
    \end{itemize}

\end{enumerate}

\end{document}